\documentclass[conference]{IEEEtran}

\usepackage{booktabs} 
\usepackage{graphicx}
\usepackage{xspace}
\usepackage[normalem]{ulem}
\useunder{\uline}{\ul}{}
\usepackage{multirow}
\usepackage{lscape}
\usepackage{rotating}
\usepackage{supertabular}
\usepackage{longtable}
\usepackage{enumerate}
\usepackage{comment}
\usepackage{ragged2e}
\usepackage{array}
\PassOptionsToPackage{hyphens}{url}
\usepackage{url}
\usepackage{hyperref}
\usepackage{array}
\usepackage{ragged2e}
\usepackage{caption}
\newcolumntype{P}[1]{>{\RaggedRight\hspace{0pt}}p{#1}}
\usepackage{fontawesome}
\usepackage[table,xcdraw]{xcolor}
\usepackage{tcolorbox}
\usepackage[switch]{lineno}
\usepackage{amsmath}

\newif\ifdraft
\drafttrue
\ifdraft
  \newcommand{\joarder}[1]{{\color{blue}\emph{Joarder: #1}}\xspace}
  \newcommand{\saleh}[1]{{\color{red}\emph{Saleh: #1}}\xspace}
\else
  \usepackage[disable]{todonotes}
  \newcommand{\joarder}[1]{}
  \newcommand{\saleh}[1]{}
\fi

\ifCLASSINFOpdf
\else
\fi
\begin{document}
%
\title{A Question Bank to Assess AI Inclusivity:  Mapping out the Journey from Diversity Errors to Inclusion Excellence}


\author{\IEEEauthorblockN{Rifat Ara Shams}
\IEEEauthorblockA{CSIRO's Data61, Australia\\
Email: rifat.shams@data61.csiro.au}
\and
\IEEEauthorblockN{Didar Zowghi}
\IEEEauthorblockA{CSIRO's Data61, Australia\\
Email: didar.zowghi@data61.csiro.au}
\and
\IEEEauthorblockN{Muneera Bano}
\IEEEauthorblockA{CSIRO's Data61, Australia\\
Email: muneera.bano@data61.csiro.au}

}


%


\maketitle

\begin{abstract}
Ensuring diversity and inclusion (D\&I) in artificial intelligence (AI) is crucial for mitigating biases and promoting equitable decision-making. However, existing AI risk assessment frameworks often overlook inclusivity, lacking standardized tools to measure an AI system's alignment with D\&I principles. This paper introduces a structured AI inclusivity question bank, a comprehensive set of 253 questions designed to evaluate AI inclusivity across five pillars: Humans, Data, Process, System, and Governance. The development of the question bank involved an iterative, multi-source approach, incorporating insights from literature reviews, D\&I guidelines, Responsible AI frameworks, and a simulated user study. The simulated evaluation, conducted with 70 AI-generated personas related to different AI jobs, assessed the question bank's relevance and effectiveness for AI inclusivity across diverse roles and application domains. The findings highlight the importance of integrating D\&I principles into AI development workflows and governance structures. The question bank provides an actionable tool for researchers, practitioners, and policymakers to systematically assess and enhance the inclusivity of AI systems, paving the way for more equitable and responsible AI technologies.
\end{abstract}

\textbf{keywords: Diversity, Inclusion, Artificial Intelligence, Question Bank}

%
\IEEEpeerreviewmaketitle

\section{Introduction}
\label{sec:intro}

Recent research highlights the critical role of diversity and inclusion (D\&I) in the development and deployment of artificial intelligence (AI), emphasizing that neglecting these considerations can lead to systemic discrimination, algorithmic oppression, and a loss of public trust \cite{shams2023ai}. AI systems, when developed without an appropriate level of D\&I focus, risk reinforcing existing societal biases, disproportionately impacting marginalized communities and exacerbating inequalities. Ensuring inclusivity in AI is not just an ethical imperative but also a necessity for creating fair, effective, and widely beneficial AI technologies \cite{avellan2020ai, bano2025does}.

In high-stakes domains such as healthcare \cite{jiang2017artificial} and recruitment \cite{bano2024diversity}, the implications of AI bias are particularly concerning \cite{chinta2024ai, gupta2022ethical}. AI-driven applications in these areas highlight the urgent need for continuous D\&I assessments to prevent biases from being embedded into decision-making processes \cite{darwish2024diversity}. Without proactive intervention, AI could inadvertently perpetuate systemic inequities, limiting opportunities and negatively affecting the well-being of diverse populations. Addressing these risks requires integrating D\&I principles at every stage of AI development, from data collection to model training and deployment \cite{bano2023ai}.

To tackle these challenges, researchers advocate for embedding D\&I considerations into AI design and governance frameworks. One approach involves utilizing tailored user story templates to explicitly capture D\&I requirements, ensuring that inclusivity is a foundational element rather than an afterthought \cite{bano2023ai}. Additionally, AI itself can serve as a powerful tool for advancing D\&I initiatives by offering data-driven insights, automating bias detection, and providing personalized interventions to promote fairness \cite{mariyono2024exploring}. For instance, IBM’s AI Fairness 360 toolkit has been used to mitigate bias in hiring algorithms \cite{bellamy2019ai}. A recent study claimed that AI guidelines focus on fairness and non-discrimination, but lack holistic approaches to promote diversity, equity, and inclusion in AI development \cite{cachat2023diversity}. Zowghi et al. addressed this gap by proposing 46 guidelines designed to enhance D\&I in AI development and deployment \cite{zowghi2023diversity}. Studies have identified many challenges and solutions for embedding D\&I principles in AI systems, emphasizing the need for ethical frameworks and diverse representation in AI development \cite{shams2023ai}. One of the key challenges in promoting D\&I in AI is the absence of standardized methods/tools/techniques for measuring or assessing an AI system's inclusivity.

A few recent studies proposed question banks and checklists as structured tools to assess responsible and explainable AI. Liao et al. introduced an extended explainable AI (XAI) question bank to represent user needs for explainability as prototypical questions \cite{liao2020questioning}. This question bank serves as a tool for creating user-centered XAI applications and informing design practices. An extended version of this study proposed a question-driven design process for XAI user experiences, grounding user needs, XAI technique selection, design, and evaluation in user questions \cite{liao2021question}. Two recent studies by Lee et al. \cite{lee2023qb4aira, lee2024responsible} introduced QB4AIRA and the Responsible AI (RAI) Question Bank, both offering structured approaches to identify and evaluate AI risks across various ethical principles. These tools aim to facilitate effective risk management and promote RAI practices. In the banking sector, Ratzan et al. developed a valid and reliable instrument to measure RAI maturity in firms, addressing the need for standardized assessment tools in this industry \cite{ratzan2023measuring}. However, none of these studies proposed any tools/techniques to assess the inclusivity of an AI system. 

Our research addresses this critical gap by introducing a structured and comprehensive approach to assessing AI inclusivity. After a rigorous analysis of literature, D\&I guidelines, Responsible AI frameworks, and a simulated user study, we proposed a systematic question bank of 253 questions to assess the inclusivity of AI systems. This approach not only enhances transparency in AI decision-making but also promotes the development of more equitable, inclusive, and socially RAI systems. Our work also contributes to the broader efforts of integrating D\&I considerations into AI governance, ensuring that AI technologies serve diverse populations fairly and effectively. The key contributions of this research are:

\begin{itemize}
    \item Development of the Inclusive AI Question Bank with 253 structured questions, categorized into five key pillars: Humans, Data, Process, System, and Governance to assist in assessing Inclusive AI.
    \item   A systematic methodology including an iterative, multi-source approach, incorporating insights from D\&I guidelines, Responsible AI frameworks, literature reviews, and AI-generated question prompts. The question bank is validated through a simulated user study with 70 AI-generated personas from diverse roles and industries to ensure broad applicability.
    \item Promoting awareness on diversity and inclusion in AI, helping stakeholders understand and integrate inclusive AI practices.
\end{itemize}

\textbf{Paper Organization.} Section \ref{sec:background} describes the background of this research and the related work. Section \ref{sec:methodology}, briefly explains our research method, and Section \ref{sec:user_study} represents the simulated user study. Section \ref{sec:results} reports the findings of this study and we discuss the findings in Section \ref{sec:discussions}. Section \ref{sec:ttv} discusses the possible threats to validity of this research. Finally, the research is concluded with possible future research directions in Section \ref{sec:conclusions}.
\section{Background and Related Work}
\label{sec:background}

\subsection{Diversity and Inclusion in AI}

Recent research has focused on defining diversity and inclusion (D\&I) in artificial intelligence (AI) and developing practical guidelines for its implementation \cite{zowghi2023diversity}. According to Zowghi et al., ``diversity and inclusion in Artificial Intelligence refers to the `inclusion' of humans with `diverse' attributes and perspectives in the data, process, system, and governance of the AI ecosystem'' \cite{zowghi2023diversity}. In this definition, attributes refer to known facets of diversity such as race, color, sex, language, religion, political or other opinion, national or social origin, property, birth or other status, age, disability, criminal record, ethnic origin, gender identity, immigrant status, intersex status, neurodiversity, sexual orientation, and intersections of these attributes \cite{zowghi2023diversity}.

The growing importance of D\&I in AI is increasingly recognized, as failing to address these critical aspects can perpetuate bias, discrimination, and even algorithmic oppression \cite{shams2023ai}. AI technologies hold immense potential to advance D\&I initiatives by revolutionizing key areas such as recruitment, personalized marketing, and the development of inclusive digital solutions \cite{mariyono2024exploring}. For instance, in talent acquisition, AI tools can help build more diverse workforces by identifying and mitigating conscious and unconscious biases during hiring processes, thereby fostering equity and inclusion in workplaces \cite{jora2022role}.

However, the journey toward achieving true inclusive AI faces challenges. Bias embedded in algorithms, coupled with a lack of diversity in training datasets, continues to undermine the equitable application of AI technologies \cite{mariyono2024exploring, bano2025does}. These limitations emphasize the need for a multi-faceted approach to address systemic issues and ensure AI operates as a force for good rather than reinforcing societal inequities. The use of AI in healthcare and recruitment illustrates the urgency of continuously assessing diversity, equity, and inclusion (DEI) impacts to safeguard against potential harm and maximize benefits \cite{darwish2024diversity}.

To advance D\&I in AI, several actionable strategies have been proposed. These include ensuring diverse datasets, implementing continuous monitoring and evaluation of AI systems, and engaging stakeholders across various demographics to reflect broader perspectives. Moreover, integrating DEI principles into ethical AI frameworks and regulatory guidelines can help align AI development with social justice goals. Ongoing research and interdisciplinary collaboration remain vital to addressing gaps and fostering innovations that support an inclusive AI ecosystem \cite{darwish2024diversity}.

Recent studies show a growing focus on D\&I in AI research, exploring how inclusive practices can help reduce biases, prevent algorithmic discrimination, and ensure fair outcomes in different applications. Zowghi et al. proposed practical guidelines for ensuring D\&I considerations are embedded within AI systems and the broader AI ecosystem \cite{zowghi2023diversity}. A recent study conducted a systematic literature review to identify a list of challenges and potential solutions to address D\&I in AI as well as using AI to enhance D\&I practices \cite{shams2023ai}. Another study explored the relationship between diversity, inclusion, and AI in multicultural workplaces and markets, highlighting the potential of AI to improve recruitment processes, personalize marketing strategies, and develop inclusive technologies, while also addressing challenges such as bias in AI algorithms and limited data diversity \cite{mariyono2024exploring}. Similarly, Jora et al. showed how AI can be used to promote diversity, equality, and inclusion in the hiring process by eliminating conscious and unconscious bias \cite{jora2022role}.

Another recent paper examined the role of diversity, equity, and inclusion (DEI) in the use of AI within the U.S. Department of Defense, highlighting its potential to enhance healthcare and recruitment while cautioning against biases in training data that could harm accuracy, trust, and equity \cite{darwish2024diversity}. Rathore et al. explored challenges in workforce diversity, biases from human and machine interactions, and how AI can help address these biases to promote D\&I \cite{rathore2022exploratory}. A recent research explored D\&I issues in AI systems by analyzing AI incident databases and found that nearly half of the incidents are related to racial, gender, and age discrimination. This study also proposed a decision tree and public repository to promote responsible and inclusive AI practices \cite{shams2024ai}. Bano et al. explored ways to identify and incorporate D\&I requirements into AI systems \cite{bano2023ai}. Another study emphasized that accountability and diversity are essential to ensuring inclusion, fairness, and bias mitigation in both AI-driven and human-controlled decision-making systems \cite{porayska2019accountability}. A recent paper critiques AI ethics guidelines for their limited focus on technical compliance with fairness and non-discrimination, advocating for practices that actively influence AI developers' behaviors and address DEI risks more comprehensively \cite{cachat2023diversity}.

To the best of our knowledge, no study has proposed any tools or techniques to assess the inclusivity of AI systems. In this study, we developed a question bank from a D\&I perspective to evaluate the inclusivity of AI systems.

\subsection{Question Bank}

Recent studies have introduced question banks and checklists as structured tools to promote responsible and explainable AI. Liao et al. presented an extended explainable AI (XAI) question bank and discussed how it can be used to support the needs specification work for creating user-centered XAI applications \cite{liao2020questioning}. The same research team proposed a question-driven design process for creating XAI user experiences that grounds the design in the types of questions users ask to understand the AI system \cite{liao2021question}. Sipos et al. assessed the XAI Question Bank's usefulness in identifying explanation needs for art historians using an AI image retrieval system and proposed an extension with 11 new questions and expanded descriptions \cite{sipos2023identifying}. QB4AIRA, a novel AI risk assessment question bank, refined questions from global frameworks, aligns them with Australia's AI ethics principles, and provides 293 prioritized questions to support stakeholders in assessing and managing AI risks \cite{lee2023qb4aira}. The extended version of this study described QB4AIRA in detail as a comprehensive AI risk assessment tool that integrates diverse AI ethics principles, offering a structured, multi-layered approach to help stakeholders at all levels identify and mitigate risks across AI projects \cite{lee2024qb4aira}. Madaio et al. described an iterative co-design process with 48 practitioners to understand the role of checklists in operationalizing AI ethics principles, specifically around the concept of fairness, and to co-design an AI fairness checklist \cite{madaio2020co}. A recent article provided a checklist of recommendations to mitigate bias in the development and implementation of AI algorithms in healthcare \cite{nazer2023bias}.

To the best of our knowledge, no existing study has proposed a structured question bank or checklist specifically designed to assess the inclusive AI systems. While various efforts have been made to develop ethical AI guidelines and fairness metrics, there remains a significant gap in standardized tools that systematically evaluate how well AI systems align with D\&I principles. This lack of assessment frameworks makes it challenging for developers, policymakers, and organizations to evaluate and improve the inclusivity of AI technologies in a consistent and actionable manner. In this regard, our research is unique, as it aims to fill this void by providing a systematic approach.
\section{Methodology}
\label{sec:methodology}
  
\begin{figure*}[!htbp]
            \centering
            \includegraphics[width=0.85\textwidth]{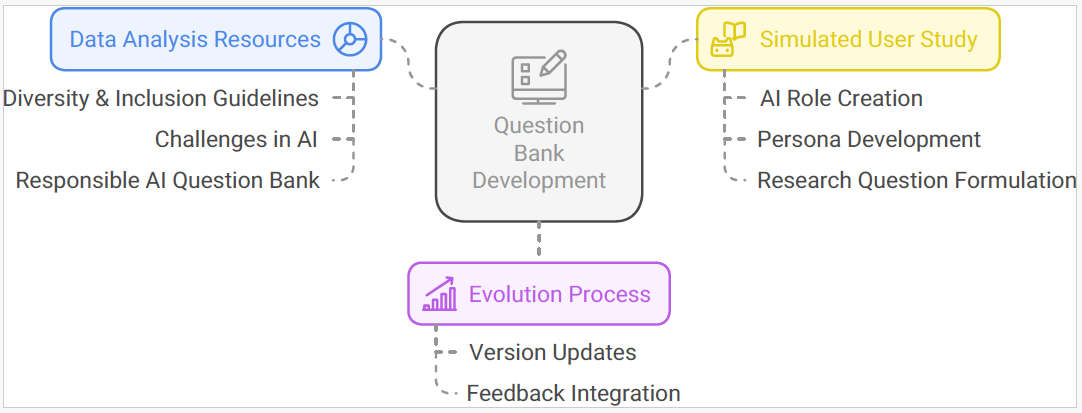}
            \caption{An overview of the research method}
            \label{fig:method}
    \end{figure*}

The research employed an iterative and collaborative methodology to develop a comprehensive Question Bank (QB) for assessing inclusivity in AI systems. The QB underwent a structured evolution through eight distinct versions, integrating insights from D\&I and RAI resources. It incorporated iterative feedback and applied triangulation techniques to ensure its reliability, relevance, and comprehensiveness (see \autoref{fig:method}).

\subsection{Resources Used for QB Development}
\textbf{Guidelines to Address D\&I in AI.} The data collection process was a foundational step in the development of the question bank designed to assess inclusivity in AI systems. The process began with a thorough review of the existing 46 D\&I guidelines relevant to AI system design, development, and deployment \cite{zowghi2023diversity}. These guidelines served as a primary reference for developing our QB by outlining best practices for inclusive AI development. They also helped identify five key pillars (humans, data, process, system, and governance) that require scrutiny in AI. As these guidelines were later updated, with Version 2 released online, we refined our QB accordingly. Specifically, 11 updated guidelines were incorporated into later versions of the QB to ensure alignment with the latest advancements in D\&I guidelines.

\textbf{Challenges to Address D\&I in AI and AI for D\&I.} As an additional data source, we utilized findings from a recent systematic literature review (SLR) to identify key challenges in two areas: addressing D\&I in AI and leveraging AI to enhance D\&I practices \cite{shams2023ai}. The study identified 55 challenges related to D\&I in AI and 24 challenges in using AI to improve D\&I efforts. This process involved analyzing all the relevant academic articles from 2017 to 2022 to capture the full scope of challenges facing the implementation of D\&I principles in AI systems. Building on insights from this SLR, we developed additional questions for our QB based on the challenges identified in the review. 

\textbf{Responsible AI (RAI) Question Bank.} We integrated questions from an RAI question bank \cite{lee2024responsible}. This RAI QB incorporated five well-known AI frameworks: NIST AI Risk Management Framework \cite{NIST}, Microsoft RAI Impact Assessment Template and Guide \cite{microsoft_template, microsoft_guide}, EU Assessment List for Trustworthy Artificial Intelligence \cite{eu_commission}, Canada Algorithmic Impact Assessment \cite{canada_impact_assessment}, and Australia's NSW AI Assurance Framework \cite{nsw_assurance}. This RAI QB was developed through a rigorous process after conducting a systematic literature review and a case study. 
This RAI QB was particularly important as it allowed for cross-validation of our QB, particularly in areas concerning diversity, inclusion, bias, and fairness. Additionally, it allowed us to incorporate questions derived from a diverse range of sources.  

\textbf{Use of a Large Language Model.} Throughout the data collection phase, GPT-4o was used as an assistant tool to generate additional questions based on consistent and predefined prompts. We used ChatGPT, as it emerged as a leading conversational AI model that outperforms other language models in generating human-like responses \cite{ahmed2023chatgpt}. We chose GPT-4o as it was the most updated version of ChatGPT while we conducted this study. These prompts were designed to produce questions that aligned closely with the identified pillars (humans, data, process, system, governance) from the D\&I guidelines and the SLR findings. The use of GPT ensured that a wide variety of questions were generated efficiently, with a focus on minimizing human bias in the question formation process. However, we ensured human-in-the-loop in each step. For example, all the authors of this article reviewed and validated all questions generated by GPT-4o. This manual validation ensured that the questions not only adhered to the guidelines but were also applicable in real-world AI contexts, free from ambiguity and potential bias.


\subsection{Evolution of the Question Bank}
\begin{figure*}[!htbp]
            \centering
            \includegraphics[width=0.9\textwidth]{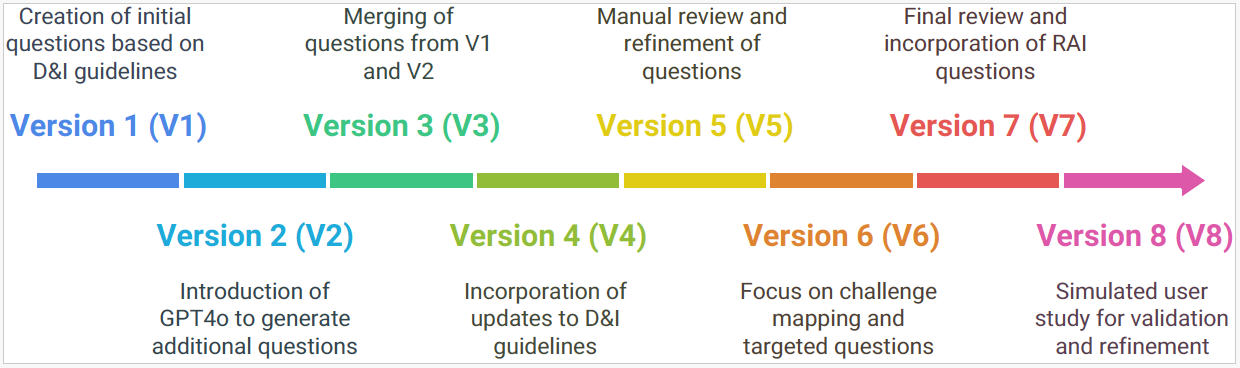}
            \caption{Question bank evolution}
            \label{fig:QB_Evolution}
    \end{figure*}

The development of the QB was an iterative process, with each version building on the insights and improvements of the previous one. The evolution of the QB occurred over eight distinct versions, each representing a major step forward in refining and expanding the set of questions designed to assess inclusivity in AI systems (see \autoref{fig:QB_Evolution}).

\textbf{Version 1 (V1).} V1 of the question bank marked the initial phase of our development process to assess the inclusivity of AI systems, where we manually developed the first set of questions based on the initial D\&I guidelines \cite{zowghi2023diversity}. To ensure comprehensive evaluation, we designed the questions to align with the five pillars that allows for a structured and systematic assessment of AI inclusivity. For each guideline within these pillars, we formulated multiple questions, typically ranging from 4 to 10 per guideline. The initial set of questions was relatively straightforward, serving as a fundamental framework upon which future iterations could build. In total, we developed 128 questions in this initial version.

\textbf{Version 2 (V2).} In V2 of the question bank, we leveraged GPT-4o to independently generate questions based on the D\&I guidelines \cite{zowghi2023diversity}. To ensure alignment with the existing 5-pillars framework, we designed a systematic approach. For each of the 46 guidelines, we provided GPT-4o with a prompt\footnote{GPT-4o prompt: Consider you are doing a survey with an AI practitioner to understand how to use the following guidelines in AI lifecycle. Create a question bank on the following guidelines for the survey. Maximum five questions please.\\
``Title of the guideline"\\ 
``Description of the guideline"} specifying the guideline and instructed it to generate a set of relevant questions aimed at assessing the inclusivity of AI systems. This process was repeated iteratively for all guidelines. Finally, GPT-4o generated a total of 230 questions. This version introduced a more systematic and diverse set of questions that created an opportunity to compare them with the manually created questions in V1.

\textbf{Version 3 (V3).} V3 compared and merged the questions from V1 and V2, removing any redundancies and ensuring that the content from both versions was harmonized into a cohesive and comprehensive set. This merging process was essential to avoid overlap and repetitions. The whole process was carried out by the first author and reviewed by the other authors and we ended up with a total of 264 questions.

\textbf{Version 4 (V4).} During the development of our question bank, the second version of the D\&I guidelines was released. 11 guidelines were updated in this version. To ensure alignment with these revisions, we leveraged GPT-4o once again to generate new questions specifically for the updated guidelines. These newly generated questions were carefully reviewed, compared, and integrated with the existing questions, ensuring that it remained fully up to date and aligned with the latest D\&I best practices. The total number of questions remained unchanged. Only linguistic refinements were made to improve clarity and alignment with the updated guidelines.

\textbf{Version 5 (V5).} V5 represented a critical refinement stage. Two of the authors manually reviewed all the questions and removed duplicates and rephrased certain questions for clarity. We also eliminated any opinion-based or subjective questions that had been identified in previous versions (e.g., Do you believe in the importance of establishing an inclusive AI ecosystem that involves the broadest range of community members?). This version was particularly focused on enhancing the clarity and usability of the QB, ensuring that each question is understood in the same way by all users. At the end of this stage, the number of questions were reduced to 223.

\textbf{Version 6 (V6).} In V6, our primary objective was to expand the scope by incorporating insights from an additional data source. This new source consisted of a comprehensive list of challenges related to D\&I in AI, as well as the ways AI can be leveraged to enhance D\&I practices \cite{shams2023ai}. This study identified 55 challenges to address D\&I in AI and 24 challenges to use AI for D\&I practices. These challenges were helpful in developing questions for our QB because they highlight real-world obstacles and gaps in current AI practices related to D\&I. By systematically identifying these challenges, we gained a clearer understanding of where AI systems may fall short in addressing D\&I concerns or perpetuate bias and exclusion. Each challenge pointed to a specific issue that may need to be examined, measured, or monitored during the design, development, or deployment of AI systems. Translating these challenges into questions allowed us to create a more comprehensive and meaningful set of questions, ensuring that the QB addresses not only high-level ethical principles, but also concrete, actionable aspects of AI inclusivity.

To systematically integrate these insights, we once again employed GPT-4o to generate a set of questions based on the identified challenges, following the same structured approach used in V2. This process resulted in 395 newly generated questions. To ensure quality and relevance, we conducted a thorough manual review of all generated questions, assessing their alignment with our existing question bank. Any questions deemed irrelevant were removed, while the remaining questions were carefully compared against V5 of our question bank. After this rigorous evaluation, we could incorporate 15 additional questions derived from the challenges into the question bank, bringing the total number of questions to 238. This step was undertaken to ensure data triangulation, enhancing the robustness and comprehensiveness of our assessment framework.

\textbf{Version 7 (V7).} In V7 we expanded our dataset by incorporating insights from an additional source, the RAI question bank \cite{lee2024responsible}. This external resource contained a total of 245 questions, covering various aspects of RAI development. To ensure alignment with our focus on D\&I, we conducted a meticulous manual analysis of all 245 questions and identified 16 questions related to bias, and fairness. Next, all the authors conducted a rigorous evaluation of these 16 questions to assess their accuracy, relevance, and consistency with the purpose of our question bank. During this review, we identified one question that did not adequately align with our framework and was therefore excluded. The remaining 15 questions were integrated into our question bank, bringing the total number of questions to 253.

\textbf{Version 8 (V8).} A simulated user study was conducted using GPT-4o to systematically validate our question bank and assess its relevance, clarity, and effectiveness across diverse AI-related roles and domains (See \autoref{sec:user_study}). This approach allowed us to gather simulated feedback from 70 personas. The insights gained from this simulated study enabled us to enhance question clarity where necessary to ensure better alignment with real-world AI challenges. As a result, we refined the language in seven questions to enhance clarity and precision. This process led to the development of V8.
\section{Simulated User Study}
\label{sec:user_study}

A simulated user study could be a powerful and cost-effective method for validating a question bank, ensuring that the questions are clear, relevant, and effective in assessing the intended knowledge or skills before deploying them in real-world settings. Lin proposed a simulation-based methodology for evaluating answers to question series, considering contextual dependencies and user behaviors \cite{lin2007user}. Breuer et al. validated simulated user query variants against real user queries, introducing a method that better reproduces real queries compared to established techniques \cite{breuer2022validating}. Ghosh et al. emphasized the importance of question bank design models and deployment strategies, suggesting intermediate checks to ensure quality before final deployment \cite{ghosh2012design}. These studies collectively demonstrate that simulated user studies can effectively validate question banks and improve the overall quality of assessments while providing insights into user behavior and query patterns. By leveraging simulated users, researchers and designers can preemptively identify ambiguities, biases, and inefficiencies in question formulation without the need for large-scale human participation, which can be expensive and time-consuming. This approach allows for iterative refinement, where questions can be tested, adjusted, and re-tested rapidly based on simulated responses, leading to higher-quality assessments.

In this research, we conducted a simulated user study using GPT-4o to validate our question bank and make potential improvements. We followed the following steps in this study.

\textbf{Step 1: AI Role Creation.} In our approach, we first used GPT-4o to generate a list of AI-related roles, initially identifying 25 roles. However, upon careful review and search, we found that some roles, such as diversity and inclusion data scientist, RAI engineer, AI fairness specialist, ethical AI program manager, were non-existent in real-world applications. To maintain accuracy and credibility, we refined the list down to 14 key roles, including data scientist, machine learning engineer, AI researcher, data engineer, AI product manager, AI/ML architect, data analyst, software engineer, AI ethicist, business analyst, project manager, user experience (UX) designer, AI policy advisor, and AI risk and compliance officer.

\textbf{Step 2: Persona Creation.} Once the roles were finalized, we proceeded to persona creation using GPT-4o. We structured the prompts\footnote{GPT-4o prompt: Create five personas on ``specific role (e.g., data scientist)" working on AI project. Each persona should have diversity in gender, race, ethnicity, age, seniority/expertise level. Create a table on these five personas mentioning their name, role, seniority/expertise level, affiliation, gender, age, race, ethnicity, size of the organization (no. of employees), domain of the organization and description of the personality.} to ensure diversity across gender, race, ethnicity, age, seniority, and domain expertise. To achieve this, we created a wide range of personas representing various professional levels: entry-level, mid-level, and senior roles, while also accounting for gender diversity. Additionally, we incorporated racial diversity by including personas from backgrounds such as Asian, Latino, Middle Eastern, and more. Similarly, we considered diversity in ethnicity, age, and domain expertise that ensured a well-rounded and inclusive representation across different demographic and professional spectrums. For each of the 14 roles, we generated five unique personas, resulting in a total of 70 personas \cite{replpack02}. These personas were documented in a structured table, capturing details such as name, role, seniority/expertise level, affiliation, gender, age, race, ethnicity, organization size (number of employees), organizational domain, and personality description.

\textbf{Step 3: Formulation of Research Questions.} To ensure a rigorous and comprehensive validation of our question bank, we formulated five research questions designed to assess the relevance, usefulness, impact, and validity of the questions. Since a standard participatory study was not conducted, we designed this simulated user study through a rigorous, iterative process involving multiple rounds of discussion among all the authors. The development of these research questions was also an iterative and collaborative process, involving several rounds of discussions. This approach ensures that the selected questions effectively capture key dimensions of AI inclusivity, aligning with diverse professional roles, application domains, and broader awareness of diversity and inclusion. Additionally, the questions facilitate critical feedback, allowing for the continuous refinement and validation of the question bank. The research questions for this simulated user study are given below.

   \noindent \textit{\textbf{Q1.} Which of the questions are relevant to your role and tasks that you carry out in your role and why do you think so?}

   \noindent \textit{\textbf{Q2.} Which of the questions are relevant to the application domain that your organization is working on? why do you think so?}

   \noindent \textit{\textbf{Q3.} What are your thoughts about the usefulness of the question bank in helping to create inclusive AI?}

   \noindent \textit{\textbf{Q4.} Are these questions helpful in educating and bringing awareness about diversity and inclusion (D\&I) in AI considerations? Explain your answer.}

   \noindent \textit{\textbf{Q5.} Would you like to add or remove or modify any questions from the question bank? If yes, which questions are those and why?}

By addressing these research questions, we aimed to gain holistic insights into the usefulness of the question bank from multiple angles including role relevance, domain applicability, inclusivity, educational value, and areas for refinement. 

\textbf{Step 4: Review of the Question Bank.} For each of the 70 personas, we used GPT-4o to generate responses to the five research questions using our question bank as the reference \cite{replpack02}. We conducted a manual review of the responses generated by GPT-4o to ensure their relevance, coherence, and alignment with the intended objectives of the study. This review process was essential to verify that the AI-generated answers did not contain irrelevant or inconsistent responses. Finally, the responses were carefully captured and stored in a structured spreadsheet, allowing for quantitative and qualitative analysis.

\textbf{Step 5: Analysis of the Review.} We conducted a thorough manual analysis of the responses generated for the five research questions for the 70 personas. The first author carefully examined each persona's feedback to extract meaningful insights and identify patterns. The findings were then documented and shared with all co-authors for collective review. To ensure a rigorous and well-rounded evaluation, all authors reviewed both the raw responses and the initial findings derived from the analysis. Following this multi-stage review process, we engaged in several rounds of discussions to discuss, refine, and consolidate our final findings.

The findings for Q1–Q4, which focused on role relevance, domain applicability, inclusivity, and educational value, are presented in Section \ref{sec:discussions} of this paper. The responses to Q5, which invited feedback on modifying, adding, or removing questions, played a crucial role in iteratively refining the question bank. By synthesizing the insights from all 70 personas, we carefully adjusted wording, removed redundant or unclear questions, and introduced new ones to better address the identified gaps. After incorporating these refinements, we arrived at the final version (V9) of our question bank, ensuring that it was comprehensive, inclusive, well-calibrated, and reflective of real-world AI challenges and considerations.
\section{Results}
\label{sec:results}
\begin{figure}[!htbp]
            \centering
            \includegraphics[width=0.45\textwidth]{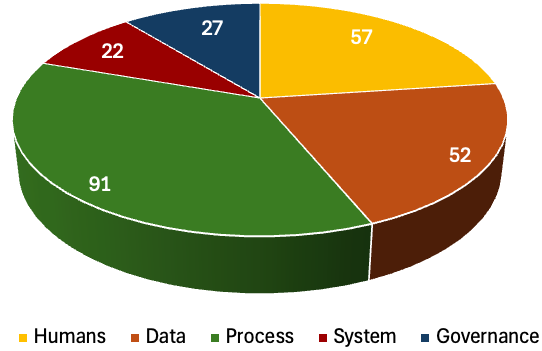}
            \caption{Number of questions in each pillar}
            \label{fig:no_of_questions}
    \end{figure}

The key contribution of this study is the development of a Question Bank aimed at assessing inclusivity in AI systems. This QB provides structured and comprehensive questions tailored to evaluate different aspects of inclusivity, encompassing human-centered design, ethical considerations, governance structures, and system-wide implementation. The Question Bank is organized in alignment with the five pillars of AI inclusivity \cite{zowghi2023diversity}. Each pillar includes a set of questions designed to probe critical inclusive AI concerns. \autoref{fig:no_of_questions} shows the number of questions in each pillar.

\subsection{Human-Centered Questions}
Human-centered questions are a critical component of the Question Bank, ensuring that AI systems are designed, developed, and deployed with a focus on diversity, equity, and inclusion. The aim is to ensure that AI-driven solutions are not only technically robust but also socially responsible and beneficial for a broad spectrum of users, particularly marginalized and underrepresented groups. Indicative examples of human-centered questions are explained below:

\faQuestionCircle{} \textit{Are end-users, AI practitioners, and subject matter experts included in considering the integration of diversity and inclusion principles throughout the AI lifecycle?}\\
This question ensures that AI systems are evaluated by a diverse range of stakeholders before being deployed. AI systems often fail to meet the needs of diverse populations because they are designed without input from the people who will use them or be affected by them. Including end-users, subject matter experts, and AI developers ensures that models align with user expectations and do not unintentionally marginalize certain groups. For instance, voice recognition AI trained without input from users with speech impairments may fail to understand diverse speech patterns, leading to accessibility issues. By incorporating varied perspectives, AI teams can identify usability barriers, eliminate exclusionary design choices, and improve overall system effectiveness.

\faQuestionCircle{} \textit{Does the AI system respect human rights, diversity, and autonomy of individuals?}\\
This question requires AI teams to evaluate the ethical implications of their models. It prompts them to conduct bias audits, impact assessments, and fairness evaluations to ensure AI is aligned with global human rights standards. AI applications, especially in areas like law enforcement, hiring, and finance, must be built to prevent discriminatory outcomes, mass surveillance, and biased decision making. Without human rights protections, AI can reinforce existing power imbalances and disproportionately harm vulnerable communities.

\faQuestionCircle{} \textit{Do you ensure that AI systems accommodate a wide range of individual preferences and abilities, considering marginalized groups?}\\
AI should be able to adapt to different user needs, including those with disabilities, language barriers, or technological limitations. For example, an AI-driven customer service tool should be accessible to blind users through screen readers and provide text-based alternatives for deaf users. Developers must ensure AI models support multiple accessibility features, such as alternative input methods, multilingual options, and compatibility with assistive technologies. Additionally, this question prompts teams to test their AI systems on diverse populations, ensuring usability across various communities.

\subsection{Data-Driven Questions}
Data-driven questions help evaluate whether AI models are trained, tested, and deployed using diverse, fair, and unbiased datasets. Since AI systems heavily rely on data to make decisions, any bias in data collection, processing, or interpretation can lead to systematic discrimination, unfair outcomes, and exclusion of certain demographic groups. To address these issues, AI teams must ensure ethical data collection, representation, privacy protections, and bias mitigation throughout the AI lifecycle. The following questions are sourced from data-related inquiries.

\faQuestionCircle{} \textit{Are opt-out options offered for data collected for model training and application in your AI projects?}\\
Ethical AI development should respect the right of individuals to privacy and data autonomy. Without opt-out options, users may feel coerced into sharing sensitive information, especially in high-risk applications like credit scoring or healthcare. Ensuring clear consent mechanisms and opt-out options allows users to control their data. Organizations should adopt GDPR-compliant data collection policies \cite{barati2019developing} and ensure that AI models do not disproportionately target or track specific populations without consent.

\faQuestionCircle{} \textit{Do you explore data sovereignty from and with the perspectives of those whose data is being used in your AI projects?}\\
Data sovereignty refers to the control that individuals or communities have over their own data \cite{hummel2021data}. Many marginalized groups, including Indigenous populations, have historically faced data exploitation without proper consent or representation. Ethical AI teams must engage directly with community representatives and establish transparent data ownership agreements. This ensures that AI systems do not extract or exploit sensitive cultural, historical, or social data without explicit permissions.

\faQuestionCircle{} \textit{Do you adopt privacy-preserving synthetic datasets in cases where real datasets are too sensitive to be released publicly?}\\
In fields such as healthcare, financial services, and national security, real-world data is too sensitive to share. However, limiting data access can also hinder AI innovation. Using synthetic datasets, which retain statistical properties of real-world data while ensuring privacy, allows for RAI training without violating privacy norms.

\subsection{Process-Oriented Questions}
Process-oriented questions assess whether inclusivity is embedded at each stage of the AI development lifecycle, from ideation and design to deployment and ongoing monitoring. These questions evaluate whether inclusion is treated as a core principle rather than an afterthought and ensure that decision-making processes are structured to address potential biases and accessibility challenges early on. Examples include:

\faQuestionCircle{} \textit{Are the socio-technical implications of multiple trade-offs considered during the design stage of your AI systems?}\\
AI teams frequently prioritize performance metrics (such as accuracy) over fairness, leading to AI models that reinforce existing disparities. AI teams should actively measure and document the trade-offs between fairness, accuracy, interpretability, and transparency, ensuring that AI models balance these competing objectives.

\faQuestionCircle{} \textit{Does your validation process deal with issues like noisy labeling such as mislabeled samples in the training data?}\\
Noisy or incorrect labeling in training datasets can lead to algorithmic biases and unfair AI predictions, disproportionately affecting certain groups. AI teams should implement rigorous data validation techniques, label auditing mechanisms, and human-in-the-loop correction processes to ensure fair labeling practices.

\faQuestionCircle{} \textit{Does your organization have any mechanisms enabling continuous monitoring and improvement of diversity and inclusion considerations in your AI projects?}\\
AI fairness is not static, it evolves as society, regulations, and user demographics change. Without continuous monitoring, AI biases may worsen over time. AI teams should implement periodic fairness audits, real-world testing, and transparency dashboards to ensure ongoing assessment and refinement of inclusivity metrics.


\subsection{System-Level Questions}
These questions analyze how inclusion principles translate into the system architecture and behavior. These questions ensure AI models do not systematically disadvantage particular groups. Unlike human-centered or process-oriented questions, system-level questions examine the inner workings of AI models, including their decision logic, adaptability, transparency, and robustness across different user demographics. Examples include:

\faQuestionCircle{} \textit{Do you incorporate diverse values and cultural perspectives from multiple stakeholders and populations in your mathematical models and AI system design?}\\
Many AI models embed assumptions and biases based on dominant cultural perspectives, failing to reflect diverse worldviews. AI developers should integrate cross-cultural testing, linguistic adaptability, and intersectionality frameworks to ensure AI is inclusive and sensitive to global diversity.

\faQuestionCircle{} \textit{Do you evaluate and adjust the effectiveness of bias mitigation during model training and implementation?}\\
Bias mitigation techniques often deteriorate over time as AI systems adapt to real-world data shifts. AI teams should regularly retrain models, fine-tune bias mitigation parameters, and integrate external fairness audits to ensure long-term equity.

\faQuestionCircle{} \textit{Does the model specification include how and what sources of bias were identified?}\\
Many AI systems lack documentation on how bias mitigation efforts were conducted, making it difficult to audit, improve, or contest AI decisions. AI teams should provide comprehensive bias reports, algorithmic transparency disclosures, and structured fairness documentation to ensure AI development remains accountable and traceable.

\subsection{Governance Questions}
Governance-related questions focus on the policies, oversight mechanisms, and regulatory frameworks to assess whether an AI system is monitored, regulated, and held accountable to ensure it upholds inclusion, fairness, and ethical standards. By requiring structured policies, compliance monitoring, fairness audits, and user feedback mechanisms, these questions help organizations institutionalize inclusion as an ongoing practice. Implementing strong governance ensures that AI systems are not only inclusive upon release but continue to evolve in fairness and accountability over time. Some examples of governance-related questions are explained below.

\faQuestionCircle{} \textit{Does your organization have established policies for how biometric data and face and body images are collected and used?}\\
The use of biometric data (face scans, body images, voice recognition, and gait analysis) raises concerns regarding privacy, consent, and potential discrimination, especially for marginalized groups. AI governance must establish explicit policies governing biometric data collection, usage, and storage. Organizations should also implement user consent protocols, bias audits, and legal compliance frameworks to protect individuals from unlawful surveillance and discrimination.

\faQuestionCircle{} \textit{Have you aligned AI bias mitigation with relevant existing and emerging State and Federal legislation covering AI use in hiring, eligibility decisions, discrimination prohibitions, privacy, and unfair or deceptive practices?}\\
AI models used in hiring, finance, healthcare, and law enforcement are subject to evolving regulations on bias mitigation, privacy, and discrimination. AI governance structures should align with legal mandates to ensure compliance with anti-discrimination laws (e.g., GDPR \cite{rahat2022your}, EEOC guidelines \cite{robinson2008language}, AI Act \cite{sovrano2022metrics}). Organizations must also regularly update governance policies based on emerging AI regulations.

\faQuestionCircle{} \textit{Is there an established protocol in your organization to update AI systems in response to changes in relevant legislation or legal standards?}\\
AI models and governance policies must be flexible and adaptable to legal, ethical, and societal shifts. Without updating AI governance frameworks, outdated AI policies may fail to protect marginalized communities from bias. Organizations should implement legal compliance tracking, continuous AI audits, and automated policy updates to ensure AI governance remains aligned with current laws and ethical AI standards.

\section{Discussions}
\label{sec:discussions}
This section discusses the findings of this study.

\subsection{Insights from Question Bank}
The question bank provides a comprehensive framework for assessing inclusion in AI systems. By analyzing the structure and content of these questions, we have identified recurring patterns that inform best practices in ensuring inclusive AI systems. The following patterns emerging from this QB highlight key thematic areas crucial for assessing inclusion in AI, spanning human involvement, data governance, development processes, and systemic accountability.

\textbf{Stakeholder Engagement and Representation.} Multiple questions focus on ensuring diverse groups are represented in AI teams, decision-making processes, and system design. In software engineering, stakeholder engagement is not optional, it is fundamental to the success of any system, particularly in AI development. This is especially critical in AI systems, where decision-making mechanisms have far-reaching ethical and societal implications. Stakeholder engagement and representation in AI development require more than just assembling diverse teams. They demand meaningful inclusion at every stage of the AI lifecycle, from ideation to deployment and monitoring. A key theme in the framework is whether AI teams actively integrate D\&I principles into their workflows, ensuring that various stakeholders—such as end-users, social scientists, ethicists, and impacted communities—are consulted in shaping AI systems (Q no. 3, Q no. 5). The QB also probes whether organizations ensure that marginalized individuals involved in AI projects are not solely responsible for representing all perspectives within their communities (Q no. 17). Beyond representation, the QB highlights the importance of stakeholder feedback loops, ensuring that diverse voices continue to shape AI systems post-deployment. The QB queries whether organizations have accessible mechanisms for reporting AI-related harms (Q no. 18) and whether feedback is consistently incorporated into system updates and communicated transparently to relevant stakeholders (Q no. 20–21).

\textbf{Accuracy and Fairness of Demographic Data.} A recurring theme in the QB is to ensure the accuracy and fairness of demographic data in AI systems. The quality and inclusiveness of demographic data play a critical role in mitigating bias and ensuring equitable AI outcomes. The questions emphasize the need for organizations to systematically assess whether their datasets accurately reflect the diversity of affected communities (Q no. 66) and whether proper protocols are in place to improve representation (Q no. 67). Additionally, the question bank raises the issue of demographic gaps in data collection (Q no. 68) and asks whether organizations actively prioritize specific demographic groups to ensure quality of service (Q no. 70). This highlights the challenge of under-representation, where AI systems trained on skewed datasets may fail to generalize effectively across diverse populations. For example, in facial recognition technology, research has shown that models trained predominantly on lighter-skinned individuals exhibit higher error rates for darker-skinned subjects. The QB's focus on assessing how different demographic groups might be affected by AI outputs (Q no. 71) reinforces the importance of actively evaluating how biases in training data translate into real-world disparities.

\textbf{Bias Detection and Mitigation Strategies.} The structure of the QB focuses on the importance of proactively identifying, addressing, and mitigating biases throughout the AI lifecycle, emphasizing a multi-faceted approach that extends beyond conventional fairness metrics. The questions highlight the need for organizations to systematically evaluate bias in data collection, model training, decision-making processes, and deployment, ensuring that AI systems do not disproportionately disadvantage certain demographic groups. The QB explicitly asks whether organizations conduct demographic data accuracy reviews (Q no. 66) and establish protocols to improve it (Q no. 67), as well as whether they assess how different stakeholder groups may be affected by AI outputs (Q no. 71–72). Furthermore, it promotes the continuous monitoring of bias through fairness evaluations and audits (Q no. 161–165), ensuring that AI systems evolve in response to emerging fairness challenges. The QB also interrogates the effectiveness of bias mitigation techniques during model training and post-deployment monitoring, ensuring that fairness interventions are not merely theoretical but actively assessed for impact. For instance, it probes whether bias mitigation strategies are documented within model specifications (Q no. 143–144) and whether teams employ validation processes to address bias-related challenges, including mislabeled training data and proxy variables (Q no. 146–148). Moreover, it evaluates whether AI models are tested in real-world conditions to prevent unintended biases from emerging in deployment (Q no. 149–150).

Now consider a scenario where a multinational company implements an AI-powered hiring system to screen and rank job applicants. The organization aims to streamline its recruitment process while maintaining fairness and inclusion. However, after deployment, candidates from minority groups report a lower selection rate despite having similar qualifications. Internal analysis reveals the AI model disproportionately favors candidates from certain demographic backgrounds due to biased training data—historical hiring decisions reflecting past organizational biases. At this stage, our QB could help in bias detection and mitigation.

Demographic Data Accuracy Review (Q66, Q67):
The hiring AI team reviews the accuracy of demographic data in their training set. They find that data labels for gender and ethnicity are inconsistently applied, which skews representation.

Stakeholder Impact Assessment (Q71, Q72):
The company assesses how different groups are affected by AI-generated rankings. Findings indicate that underrepresented ethnic groups are disproportionately filtered out at the screening stage.

Bias Mitigation Documentation (Q143, Q144):
The team documents the bias mitigation strategies applied in retraining the model, ensuring future transparency in decision-making processes.

Validation and Testing with Real-World Data (Q146–Q150):
The AI model is tested in real-world hiring conditions to assess its performance across diverse candidate pools.

Continuous Fairness Monitoring (Q161–Q165):
Post-deployment, the company establishes an audit system to track fairness metrics and dynamically adjust the AI system when new biases emerge.

\textbf{Transparency and Accountability} Several questions highlight the need to monitor AI systems after deployment, maintain accountability, and establish governance to prevent unintended harm. The QB probes whether AI systems are assessed beyond the code level, considering their institutional impact, embedded values, and interaction with users and affected communities (Q no. 213–215). The QB also promotes the establishment of clear policies and procedures for handling system outputs and unexpected behaviors, ensuring that deviations from planned outcomes trigger corrective action (Q no. 183–184). Accountability is also framed as an organizational responsibility, requiring structured governance and oversight mechanisms to ensure ethical AI development. For example, the QB investigates whether AI governance policies are aligned with existing and emerging legal standards to prevent unfair or deceptive practices (Q no. 236–240). AI accountability is also linked to organizational leadership, with questions exploring whether organizations have dedicated personnel responsible for implementing and monitoring AI ethics compliance (Q no. 247–248) and whether leaders are trained to operationalize RAI practices (Q no. 249–250).

If we consider another scenario, where a large financial institution implements an AI-powered loan approval system to assess customer eligibility and streamline decision-making. The model considers factors such as income, credit history, and spending patterns. However, shortly after deployment, customers start complaining that the AI system rejects certain loan applications without clear explanations. A consumer advocacy group raises concerns about the lack of transparency in the decision-making process, especially for applicants from lower-income backgrounds. Regulators demand an audit, and the bank must demonstrate accountability for its AI-driven decisions. Therefore, our QB could be used in this case to enhance transparency, build consumer trust, and ensure regulatory compliance.

Establishing Policies for Handling System Outputs (Q183–Q184):
The institution implements clear protocols for addressing unexpected behaviors, ensuring that when deviations from planned outcomes occur (e.g., high rejection rates for certain demographics), corrective actions are triggered. A human review process is introduced for flagged loan applications to prevent unfair denials.

Ensuring Compliance with Legal Standards (Q236–Q240):
The organization aligns its AI governance policies with financial regulations and anti-discrimination laws. Regular bias audits and explainability tests are conducted to ensure the model does not violate fair lending practices.

Assigning AI Ethics Responsibility (Q247–Q248):
The bank appoints a dedicated AI ethics officer to oversee compliance, monitor for algorithmic biases, and act as a point of contact for customer concerns. A real-time monitoring dashboard is introduced to detect potential algorithmic drift, where system behavior changes over time.

\textbf{Equitable Access and AI Literacy.} Many questions address barriers to AI participation, emphasizing the role of education, community engagement, and infrastructure access in promoting a more inclusive AI landscape. Equitable access is framed as not only providing physical infrastructure—such as computing resources and network connectivity (Q no. 33)—but also ensuring that communities have the skills, knowledge, and opportunities to engage with AI systems. The question bank probes whether organizations support training and upskilling initiatives for new AI practitioners (Q no. 34), collaborate with civil society organizations to expand AI education efforts (Q no. 35), and allocate specific roles within teams to address access disparities (Q no. 36). AI literacy is another crucial factor in ensuring broader participation. The QB evaluates whether organizations implement structured AI education programs (Q no. 38), including dedicated efforts to develop, promote, and tailor AI literacy programs for specific cohorts (Q no. 39). A strong emphasis is placed on embedding diversity, equity, and bias awareness into AI training initiatives (Q no. 40–42), ensuring that AI practitioners understand the potential risks and ethical challenges in algorithmic decision-making. Additionally, the question bank assesses whether organizations provide educational resources on data biases (Q no. 43) and incorporate diversity education into employee onboarding processes (Q no. 44), reinforcing AI literacy as an ongoing learning process rather than a one-time training module.

\subsection{Insights from Simulated User Study}
\begin{table*}
\centering
\caption{Roles of the personas and the frequency of their corresponding relevant questions}
\label{table:user_study_Q1}
\resizebox{\textwidth}{!}{%
\renewcommand{\arraystretch}{1}
\begin{tabular}{|p{6cm}|p{5cm}|p{5cm}|p{5cm}|p{5cm}|p{2cm}|}

\hline
\multicolumn{1}{|c|}{\multirow{2}{*}{\textbf{Roles}}} & \multicolumn{5}{c|}{\textbf{Frequency of relevant questions under each pillar}}                                                                                             \\ \cline{2-6} 
\multicolumn{1}{|c|}{}                                & \multicolumn{1}{l|}{Humans} & \multicolumn{1}{l|}{Data} & \multicolumn{1}{l|}{Process} & \multicolumn{1}{l|}{System} & Governance \\ \hline
Data Scientist                                        & \multicolumn{1}{l|}{16}     & \multicolumn{1}{l|}{12}   & \multicolumn{1}{l|}{5}       & \multicolumn{1}{l|}{0}      & 8          \\ \hline
Machine Learning Engineer                             & \multicolumn{1}{l|}{14}     & \multicolumn{1}{l|}{16}   & \multicolumn{1}{l|}{9}       & \multicolumn{1}{l|}{0}      & 3          \\ \hline
AI Researcher                                         & \multicolumn{1}{l|}{15}     & \multicolumn{1}{l|}{12}   & \multicolumn{1}{l|}{6}       & \multicolumn{1}{l|}{2}      & 7          \\ \hline
Data Engineer                                         & \multicolumn{1}{l|}{8}      & \multicolumn{1}{l|}{15}   & \multicolumn{1}{l|}{8}       & \multicolumn{1}{l|}{0}      & 4          \\ \hline
AI Product Manager                                    & \multicolumn{1}{l|}{16}     & \multicolumn{1}{l|}{5}    & \multicolumn{1}{l|}{14}      & \multicolumn{1}{l|}{0}      & 7          \\ \hline
AI/ML Architect                                       & \multicolumn{1}{l|}{12}     & \multicolumn{1}{l|}{12}   & \multicolumn{1}{l|}{5}       & \multicolumn{1}{l|}{2}      & 10         \\ \hline
Data Analyst                                          & \multicolumn{1}{l|}{12}     & \multicolumn{1}{l|}{16}   & \multicolumn{1}{l|}{6}       & \multicolumn{1}{l|}{0}      & 4          \\ \hline
Software Engineer                                     & \multicolumn{1}{l|}{14}     & \multicolumn{1}{l|}{8}    & \multicolumn{1}{l|}{10}      & \multicolumn{1}{l|}{2}      & 4          \\ \hline
AI Ethicist                                           & \multicolumn{1}{l|}{16}     & \multicolumn{1}{l|}{3}    & \multicolumn{1}{l|}{10}      & \multicolumn{1}{l|}{0}      & 14         \\ \hline
Business Analyst                                      & \multicolumn{1}{l|}{14}     & \multicolumn{1}{l|}{3}    & \multicolumn{1}{l|}{11}      & \multicolumn{1}{l|}{2}      & 7          \\ \hline
Project Manager                                       & \multicolumn{1}{l|}{15}     & \multicolumn{1}{l|}{0}    & \multicolumn{1}{l|}{13}      & \multicolumn{1}{l|}{1}      & 10         \\ \hline
User Experience (UX) Designer                         & \multicolumn{1}{l|}{15}     & \multicolumn{1}{l|}{0}    & \multicolumn{1}{l|}{13}      & \multicolumn{1}{l|}{11}     & 0          \\ \hline
AI Policy Advisor                                     & \multicolumn{1}{l|}{12}     & \multicolumn{1}{l|}{3}    & \multicolumn{1}{l|}{10}      & \multicolumn{1}{l|}{0}      & 14         \\ \hline
AI Risk and Compliance Officer                        & \multicolumn{1}{l|}{10}     & \multicolumn{1}{l|}{6}    & \multicolumn{1}{l|}{7}       & \multicolumn{1}{l|}{0}      & 16         \\ \hline

\end{tabular}
}
\end{table*}

\textbf{Insights from Q1.}  ``Which of the questions are relevant to your role and tasks that you carry out in your role and why do you think so?''. From the responses, we manually analyzed the frequency of the relevant questions under each pillar for all the 14 roles of the personas, which is shown in \autoref{table:user_study_Q1}. The high frequency of relevant questions under the 'Humans' pillar across the majority of roles underscores the critical importance of human involvement in ensuring the inclusivity and ethical alignment of AI systems. 
This concept is essential for ensuring that AI systems align with human values, particularly regarding fairness, ethics, and inclusivity. The table shows that 10 out of the 14 roles have the highest relevance to human-related questions, which is significant because it highlights the importance of human-centric concerns across the AI lifecycle.

Another interesting observation from \autoref{table:user_study_Q1} is, that 8 out of the 14 roles have zero relevance with system-related questions. These roles might focus on areas such as human impact, data ethics, or governance, which are more concerned with the broader implications of AI, rather than the technical inner workings. For example, AI Policy Advisors may not concern themselves with how an AI model is built or trained but instead, whether its use complies with privacy laws and human rights regulations. However, while governance and human-centric concerns are important, the lack of focus on system-level questions creates a gap in developing inclusive AI. AI systems should be validated not only for their outputs but also for their architecture, data handling, and decision-making processes to avoid reinforcing biases. To address D\&I effectively, system-level guidelines must be expanded to ensure inclusivity is considered at every stage of the AI development lifecycle.

Another observation from the table is that, with the exception of the UX Designer, all other roles have a significant number of Governance-related questions. This suggests that most roles are concerned with the broader regulatory, compliance, and ethical aspects of AI systems, such as ensuring that AI adheres to laws, policies, and ethical standards. However, the UX Designer stands out because they do not seem to focus on governance-related questions (with a score of 0). This is because the primary responsibility of a UX Designer is to ensure that the AI system provides an inclusive and user-friendly experience, not necessarily to ensure compliance with governance frameworks. Their questions are more likely to address user interaction, accessibility, and usability rather than the regulatory and policy implications of the system. Similarly, except for the Project Manager, all the other roles have a good number of Data-related questions. Data is central to AI systems, as the quality and type of data used can significantly influence the performance and fairness of the system. The Project Manager's role is generally focused on the logistical and operational aspects of the project, such as coordinating timelines, resources, and team activities. While they may ensure that data is managed and used effectively, their primary concern is not the technical or ethical considerations of the data itself.


\textbf{Insights from Q2.} ``Which of the questions are relevant to the application domain that your organization is working on? why do you think so?''. We manually analyzed the responses across all domains where the personas work, and identified several common themes as critical areas of focus in the relevant questions. They are:

\begin{enumerate}
    \item Data Diversity and Bias Mitigation: Data diversity and bias mitigation are key concerns in all domains, aiming to ensure fairness in AI systems. This involves using diverse datasets and conducting fairness audits to avoid reinforcing stereotypes or discrimination, ensuring AI models perform equitably across different groups.
    \item Socio-Technical Considerations: Socio-technical considerations focus on how AI systems interact with society and human users. The goal is to design technology that is both technically effective and socially responsible, prioritizing adaptability, inclusivity, and human-centered design.
    \item AI Governance and Risk Management: AI governance and risk management are central to maintaining ethical standards and compliance. These concerns include establishing frameworks to prevent risks like algorithmic bias, and data misuse, and ensuring AI systems are aligned with legal and ethical guidelines.
    \item Stakeholder Feedback and Engagement: Stakeholder feedback and engagement highlight the importance of involving relevant groups in the AI development process. Engaging stakeholders ensures that AI systems meet societal needs, address concerns, and improve the inclusivity and effectiveness of the technology.
\end{enumerate}

\textbf{Insights from Q3.} ``What are your thoughts about the usefulness of the question bank in helping to create inclusive AI?''. We identified the most useful common areas of the question bank based on insights from 70 personas. They are:

\begin{enumerate}
    \item Structured Framework: Provides a systematic and comprehensive approach to integrating D\&I principles throughout the AI lifecycle, from design to implementation and governance.
    \item Promotes Fairness and Transparency: Encourages ethical practices, fairness, accountability, and transparency in AI systems, fostering trust and reducing bias.
    \item Guidance Across AI Lifecycle: Supports D\&I considerations at every stage of AI development, including data handling, model creation, deployment, and governance.
    \item Sector-Specific Applications: Adaptable to various fields such as healthcare, finance, sustainability, cybersecurity, and human rights, addressing unique challenges and needs.
    \item Supports Ethical and Inclusive Practices: Aligns AI development with ethical standards, stakeholder inclusivity, and community-focused goals, ensuring responsible and socially impactful AI solutions.
    \item Educational Resource: Provides valuable insights and learning opportunities for entry-level professionals, designers, and project managers to understand and apply D\&I principles effectively.
    \item Enhances Accessibility and User Trust: Encourages the development of accessible, empathetic, and user-friendly AI interfaces that prioritize diverse perspectives and build trust among stakeholders.
    \item Supports Compliance and Risk Management: Guides organizations in adhering to regulatory, privacy, and risk considerations, particularly in compliance-heavy industries.
    \item Fosters Standardized Practices: Helps large organizations maintain consistency in ethical and inclusive AI practices across teams and projects.
    \item Community and Mission Alignment: Ensures AI aligns with broader organizational missions, such as sustainability, social innovation, and equity, while addressing inclusivity gaps effectively.
\end{enumerate}


\textbf{Insights from Q4.} ``Are these questions helpful in educating and bringing awareness about diversity and inclusion (D\&I) in AI considerations? Explain your answer''). From the 70 personas, the QB has been highly effective in educating and raising awareness about D\&I in AI considerations. One common theme that emerged is the structured approach of the questions, which allows individuals, especially those new to the field, to easily understand and implement D\&I practices across the AI lifecycle. The questions provide comprehensive coverage of various D\&I aspects, such as fairness, bias mitigation, transparency, and inclusion, which resonate across multiple domains, including healthcare, finance, and social innovation.

Many responses highlighted the QB encourages proactive engagement with D\&I by prompting professionals to think critically about their AI systems' impact on diverse demographic groups. It helps teams reflect on how AI systems can be made more inclusive and ethically responsible at each stage, from data collection and model development to deployment and governance. For instance, individuals in healthcare and finance emphasized that the questions guided them in addressing biases and ensuring that their AI solutions were fair and transparent, directly contributing to better outcomes in sensitive areas like patient care and financial services.

Furthermore, the QB was perceived  as a valuable tool in raising awareness and fostering a culture of inclusion within teams, especially those with entry-level professionals. Several personas mentioned that the questions were essential in educating newcomers about the importance of D\&I, offering them a foundational understanding of how inclusion can be embedded into AI systems. This educational framework is not only about ethical implications but also about ensuring that D\&I practices are aligned with organizational goals, whether it is promoting social equity in nonprofit sectors or creating ethically sound AI models in business environments.


\subsection{Overlapping Questions in Multiple Pillars}
The questions in the QB are divided into five main pillars. These pillars help organize the questions and cover different areas of AI. However, many of the questions are not strictly confined to a single pillar, instead, they often span across multiple pillars. This is because the different aspects of AI systems, such as their design, data usage, processes, and societal impact, are inherently interconnected. For example, a question about how fair an AI model's decisions are might be related to both the data it uses and the system it runs on. We also identified several questions in the QB that falls under multiple pillars. For example, the question (Q no. 94), ``Do you have policies and protocols enabling the public release of statistical summaries and anonymized training and validation data sets (including synthetic data sets)?'' can be classified under both the data and governance pillars. Similarly, the question (Q no. 95), ``Do you assess dataset suitability factors in your AI development process?'' can fall under both the data and process pillars.

Additionally, the QB is designed to be context-dependent and use-case specific, meaning that the relevance of certain questions can change based on how the AI system is being used. For example, a question about inclusion in user interfaces might fall under the ``humans'' pillar in one case, but in another case, it could be more about the ``system'' depending on whether it is about user experience or technical features.

In this section, we have mentioned these overlaps to state that it is not uncommon for questions to appear in more than one pillar. We want QB users understand that these overlaps are part of the design and that the questions should not be strictly tied to a single pillar. Instead, they should be considered based on the specific context and purpose of the AI system. 

\subsection{Comparison with Existing AI Question Banks}
Two of the most relevant existing works are the Explainable AI (XAI) Question Bank \cite{liao2020questioning} and the AI Risk Assessment Question Bank (QB4AIRA) \cite{lee2023qb4aira}. However, their objectives differ significantly from ours, as neither was specifically designed to assess AI inclusivity. Furthermore, our question bank consists of 253 questions covering five key pillars, ensuring a holistic evaluation of AI inclusion. While QB4AIRA includes 16 questions related to bias and fairness, these do not comprehensively address D\&I which is the core focus of our QB. Additionally, 16 questions are insufficient for a thorough evaluation of inclusion in AI.

Additionally, unlike the previous works that focus primarily on technical AI risks or explainability, our QB serves as a practical tool for AI developers, policymakers, and researchers to systematically assess AI inclusivity across various domains. While other question banks emphasize algorithmic transparency or risk management, our approach goes further by incorporating a significant number of questions that examine not only technical dimensions such as data, process, and system but also broader socio-technical aspects like human impact and governance. By introducing a dedicated inclusivity assessment tool, our paper advances AI evaluation beyond explainability and risk assessment, ensuring that AI systems are not only understandable and safe but also fair and inclusive.
\section{Threats to Validity}
\label{sec:ttv}
Despite the rigorous methodology employed in the development and validation of the question bank, certain limitations may affect the reliability, generalizability, and interpretation of our findings. Below, we describe each potential threat and the mitigation strategies adopted to minimize its impact.

\subsection{Internal Validity Threats}
\textbf{Bias in Question Formulation.} The process of formulating inclusivity-related questions involved human judgment, which may have introduced unintentional biases. Although we used multiple sources to develop this QB, our interpretations could have influenced the phrasing, emphasis, or scope of certain questions. Moreover, the use of LLM (GPT-4o) for generating additional questions, while useful in diversifying content, introduces the possibility of AI-generated biases based on pre-existing training data. However, to reduce human bias, the question bank was iteratively reviewed by all the authors specializing in AI ethics, diversity, and inclusion. We ensured a human-in-the-loop validation process for AI-generated questions, where all GPT-4o-created questions were manually reviewed, reworded, or discarded if found inappropriate.

\textbf{Limited Real-World Testing.} Although the QB underwent validation through a simulated user study involving 70 AI-generated personas, it has not yet been tested in real-world AI development projects. Simulated feedback may not fully capture the complexities of actual AI deployment environments where inclusion challenges manifest differently. However, to approximate real-world use cases, we diversified personas across multiple AI roles (e.g., AI ethicist, software engineer, data scientist, policy advisor, UX designer). The simulated study was structured to include sector-specific applications as well to ensure that the QB remains relevant to multiple domains (healthcare, finance, governance, etc.). Future iterations of the AIQB will undergo empirical validation in live AI development projects, collecting real-time feedback from AI practitioners, regulators, and end-users.

\textbf{Overlap Across Question Categories.} Some questions in the QB span multiple pillars, which may introduce redundancy and complicate categorization. This could lead to difficulty in interpreting responses. Therefore, users are encouraged to apply questions flexibly based on the specific AI system or context rather than adhering strictly to pre-defined categories. Future updates will incorporate feedback-driven reorganization, ensuring the question bank remains intuitive and well-structured.


\subsection{External Validity Threats}
\textbf{Context Dependency.} The relevance of the QB varies across different AI applications, industries, and regulatory environments. A question that is highly applicable in one domain (e.g., finance or hiring) may have limited relevance in another (e.g., entertainment AI systems). Additionally, AI regulations differ across regions (e.g., EU AI Act vs. US AI frameworks), which may impact the applicability of certain inclusivity criteria. However, we ensured representation from multiple domains in the simulated study, covering healthcare, finance, sustainability, governance, cybersecurity, and human rights. Furthermore, future iterations will involve cross-sector validation by AI practitioners in diverse industries, ensuring global applicability.

\textbf{Evolving AI Ethics and Regulatory Standards.} AI ethics and Responsible AI standards are continuously evolving, therefore, some questions may become outdated or misaligned with future best practices and legal requirements. Hence, the QB will be updated periodically based on emerging AI guidelines. A feedback mechanism will also be established that would allow practitioners and researchers to contribute new questions and modifications.


\subsection{Construct Validity Threats}
\textbf{Respondent Interpretation Variability.} Different AI professionals may interpret the questions in our QB differently, that might lead to subjective variability in responses. For example, a software engineer may approach a question about bias mitigation differently than an AI policymaker. Therefore, there is a potential threat of inconsistencies in how inclusivity is assessed. However, to enhance clarity, the questions are designed with precise, non-ambiguous wording with the aim to avoid subjective phrasing. Future iterations will also incorporate empirical response analysis to refine question clarity and consistency.
\section{Conclusions}
\label{sec:conclusions}
This research introduces the inclusive AI question bank as a structured framework for assessing and improving AI inclusivity across various domains. By addressing the gap in AI risk assessment tools, the question bank provides a comprehensive approach to evaluating inclusivity at different stages of the AI development lifecycle. The findings from the simulated user study affirm its relevance and effectiveness, emphasizing the need for integrating inclusivity considerations into AI governance, data preparation, model training, and deployment strategies.

Future work will focus on real-world validation of the QB by applying it across multiple industries, including healthcare, finance, education, and recruitment, to assess its effectiveness in assessing inclusion in AI systems. We also plan to make the QB open-source, which will enable researchers, practitioners, and the general public to contribute feedback, suggest refinements, and enhance its applicability. Additionally, we plan to develop an interactive digital platform featuring a dashboard and scoring system that will allow organizations to evaluate their D\&I maturity and track improvements over time. To demonstrate the QB's practical utility, we will conduct case studies, such as evaluating how organizations use AI for recruitment and decision-making, ensuring that fairness and inclusivity principles are applied effectively. Finally, as AI ethics and regulations continue to evolve, we will continuously update the QB to align with emerging global AI governance frameworks, that will ensure its long-term relevance and impact as a standardized tool for promoting inclusive AI development.

\bibliographystyle{IEEEtran}
\bibliography{Main}

\end{document}